\documentclass[letterpaper, 10 pt, conference]{ieeeconf}  

\IEEEoverridecommandlockouts                              

\usepackage{cite}
\usepackage{xspace}

\usepackage{caption}
\usepackage{graphicx}
\usepackage[font=small]{caption}
\usepackage{amsmath}
\usepackage{amssymb}
\usepackage{xcolor}
\usepackage{hyperref}
\hypersetup{colorlinks,linkcolor={black},citecolor={green},urlcolor={gray}}

\usepackage{algorithm} 
\usepackage[noend]{algpseudocode} 

\usepackage{lipsum}
\usepackage{subcaption}

\usepackage{booktabs}

\makeatletter
\DeclareRobustCommand\onedot{\futurelet\@let@token\@onedot}
\def\@onedot{\ifx\@let@token.\else.\null\fi\xspace}

\makeatother

\definecolor{MyDarkBlue}{rgb}{0,0.08,1}
\definecolor{MyDarkGreen}{rgb}{0.02,0.6,0.02}
\definecolor{MyDarkRed}{rgb}{0.8,0.02,0.02}
\definecolor{MyDarkOrange}{rgb}{0.40,0.2,0.02}
\definecolor{MyPurple}{RGB}{111,0,255}
\definecolor{MyRed}{rgb}{1.0,0.0,0.0}
\definecolor{MyGold}{rgb}{0.75,0.6,0.12}
\definecolor{MyDarkgray}{rgb}{0.66, 0.66, 0.66}
\definecolor{MyPink}{rgb}{1, 0.75, 0.79}
\definecolor{GreenStarColor}{rgb}{0.54, 0.84, 0.41}
\definecolor{MSBlue}{rgb}{0, 0.35, 0.49}

\newcommand{\bm}[1]{\boldsymbol{#1}}

\def\OURS{MiniBEE\xspace}
\def\OURSLONG{Miniature Bimanual End-effector}
\title{\LARGE \bf \OURS: A New Form Factor for Compact Bimanual Dexterity }

\author{Sharfin Islam\authorrefmark{1}\authorrefmark{2}, Zewen Chen\authorrefmark{1}\authorrefmark{2}, Zhanpeng He\authorrefmark{1}\authorrefmark{3}, Swapneel Bhatt\authorrefmark{2}, Andres Permuy\authorrefmark{2},\\
Brock Taylor\authorrefmark{2}, James Vickery\authorrefmark{2},  Zhengbin Lu\authorrefmark{3}, Cheng Zhang\authorrefmark{3}, Pedro Piacenza\authorrefmark{2}, Matei Ciocarlie\authorrefmark{2}\\\\ 
\href{https://roamlab.github.io/minibee}{\textbf{roamlab.github.io/minibee}}
\thanks{\authorrefmark{1} joint first authorship}%
\thanks{\authorrefmark{2}Dept. of Mechanical Engineering, \authorrefmark{4}Dept. of Computer Science}%
\thanks{Columbia University, New York, NY 10027, USA}%
\thanks{Corresponding email:~\texttt{si2395@columbia.edu}}%
\thanks{This work was supported in part by the NSF under awards PFI-232975 and CMMI-2037101.}
}

\begin{document}
\maketitle
\thispagestyle{empty}
\pagestyle{empty}

\begin{abstract}

Bimanual robot manipulators can achieve impressive dexterity, but typically rely on two full six- or seven-degree-of-freedom arms so that paired grippers can coordinate effectively. This traditional framework increases system complexity and footprint while only exploiting a fraction of the overall workspace for dexterous interaction. We introduce \OURS (\OURSLONG), a compact system in which two reduced-mobility arms (3+ DOF each) are coupled into a kinematic chain that preserves full relative positioning between grippers and enables the entirety of systems workspace to be used for dexterity. To guide our design, we formulate a kinematic dexterity metric to evaluate different kinematic designs. The resulting system supports two complementary modes: (i) wearable kinesthetic data collection with self-tracked gripper poses, and (ii) deployment on a standard robot arm, extending dexterity across its entire workspace. We present kinematic analysis and design optimization methods for maximizing dexterous range, and demonstrate an end-to-end pipeline in which wearable demonstrations train imitation learning policies that perform robust, real-world bimanual manipulation.
\end{abstract}

\begin{keywords}
Bimanual manipulation, dexterous manipulation, wearable imitation learning
\end{keywords}

\section{Introduction}

In recent years, bimanual robotic manipulators have shown remarkable dexterity. The combination of imitation learning from human demonstrations and two well-articulated kinematic chains has enabled such systems to use simple parallel grippers for highly dexterous tasks~\cite{hou2019review, seita2022speedfolding, bimanualRL2019, fu2024mobile, zhao2024aloha, fang2024airexo, chi2024universalmanipulationinterfaceinthewild}, with robustness to initial conditions or perturbations encountered during execution~\cite{chi2024diffusionpolicyvisuomotorpolicy,zhu2025unifiedworldmodelscoupling,trilbmteam2025carefulexaminationlargebehavior }.

To achieve these results, current systems typically rely on the combination of two 6- or 7-degree-of-freedom (DOF) robotic arms. The \textit{dexterous workspace} of the overall system, defined here as the space where the two grippers have complete positioning ability w.r.t. each other, thus lies at the intersection of the two arms' individual workspaces. As a result, this dexterous workspace is relatively small, and only covers a fraction of each arm's overall individual workspace. Moreover, this dexterous workspace is fixed relative to the base of the entire system, meaning any object must be brought into this dexterous portion of the system's workspace to be manipulated. Overall, this traditional framework for bimanual manipulation is capable of dexterity, but has limitations in terms of efficiency, footprint, and overall mobility. 

\begin{figure}[t!]
\setlength{\tabcolsep}{0.0mm}
\centering
\includegraphics[width=\linewidth]{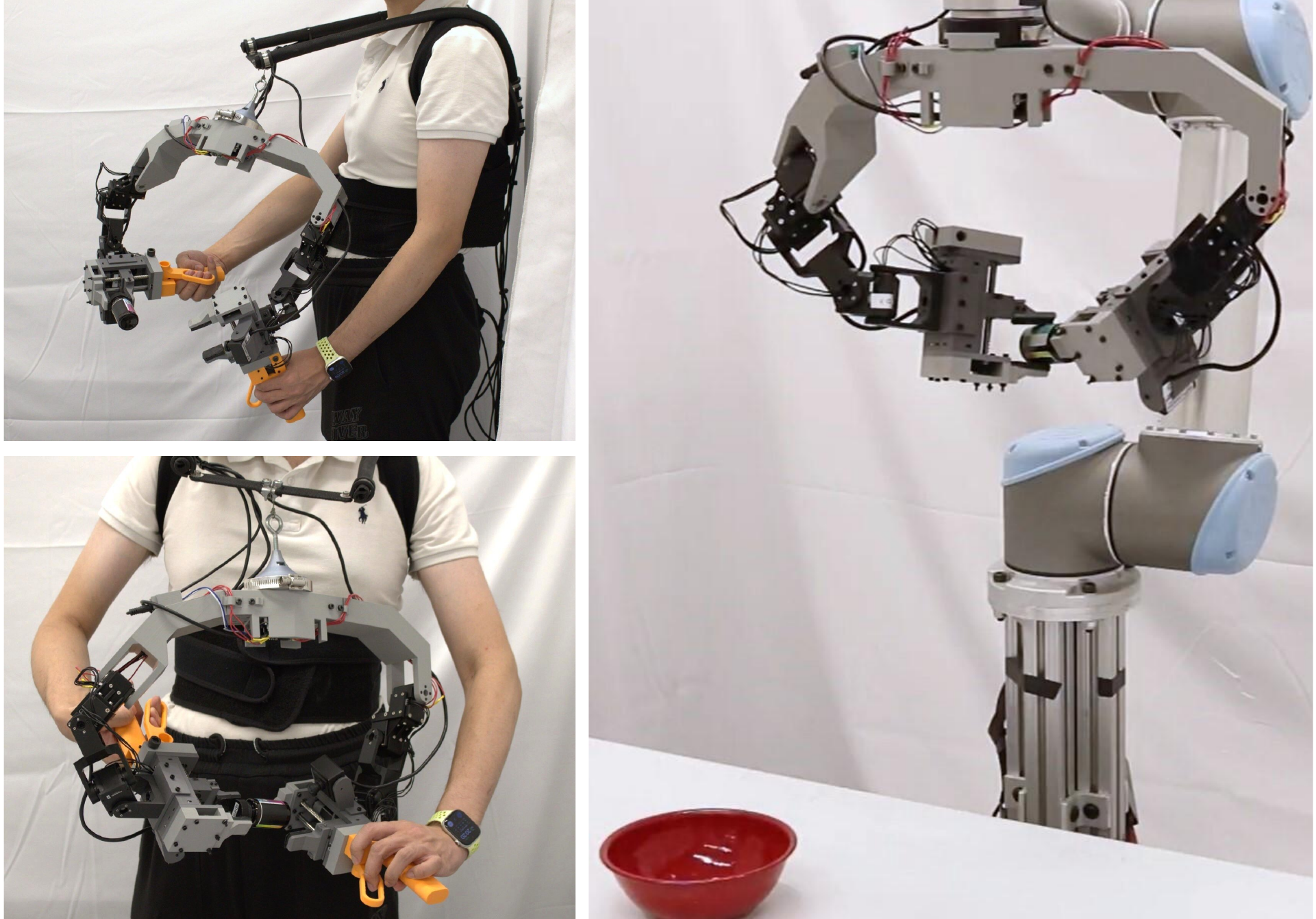}
\caption{Overview. The \textbf{\OURS} is a bimanual end-effector consisting of two compact arms equipped with grippers. The kinematic chain connecting the two grippers has sufficient degrees of freedom so that, with careful kinematic optimization, the two grippers have full manipulability w.r.t. each other, enabling high dexterity. The \OURS is compact enough to be worn and operated for kinesthetic data collection (left) for behavioral cloning, and the trained policies can be deployed with the \OURS mounted on a standard robot arm (right), giving the overall system access to almost the entirety of the overall systems workspace for dexterity.}
\label{fig:eyecandy}
\end{figure}

In this paper, we propose a new approach to bimanual dexterity that aims to address the challenges described above for bimanual systems while preserving many of their benefits. Our fundamental insight is as follows. If each of the arms of a bimanual system has reduced mobility (three+ DOF), the \textit{kinematic chain connecting the two grippers} still has sufficient articulation (6+ DOF) to allow \textit{arbitrary positioning of the grippers w.r.t. each other}, and, with careful kinematic design, we can ensure that the dexterous workspace of the overall system is large enough for bimanual tasks. We dub our system the \textbf{\OURS}, for \OURSLONG.

Thanks to the reduced articulation of each individual arm, the overall footprint of \OURS remains compact and lightweight. In turn, this provides two key benefits:
\begin{itemize}
\item The \OURS can be worn directly by an operator, allowing for easy and mobile kinesthetic data collection. When this data is used to train autonomous policies, the encoders in the kinematic chain provide full information on the relative pose of the two grippers, removing the need for external tracking or SLAM while the \OURS is manipulating an object held in or transferred between its grippers.
\item The \OURS can also be mounted on a commodity robot arm, for example when a trained policy is deployed. This means that the entire reachable area of the robot arm instantly becomes available for the \OURS, and the dexterous workspace (i.e. the area where the two grippers have complete positioning authority w.r.t. each other) is scaled by the overall reach of the robot arm.
\end{itemize}

The \OURS functions both as wearable mechanism for easy data collection with self-tracking dexterity, and as a bimanual end-effector allowing a single traditional robot arm to become dexterous over the entirety of its workspace. Overall, our main contributions are as follows. 
\begin{itemize}
\item To the best of our knowledge, this is the first example of reduced mobility arms enabling full relative mobility for bimanual tasks in a system that is compact enough for wearable collection of kinesthetic demonstrations.
\item We introduce a kinematic analysis tool that allows a designer to maximize the dexterous workspace while limiting the total number of DOF, and use these to compare multiple possible design variants. 
\item We demonstrate the end-to-end pipeline consisting of wearable kinesthetic data collection followed by training standard imitation learning policies which leverage the proprioceptive component of the demonstrations to achieve robust, dexterous bimanual manipulation with large reach in the real world.
\end{itemize}

\section{Related Work}

Bimanual systems offer several advantages for high dexterity, such as accessible off-the-shelf hardware and reliable joint control. With the appropriate kinematic structure and control policy, such systems can perform a wide range of dexterous robotic tasks. A central challenge, however, lies in teaching robots such coordination. With recent advances in learning-based control, success in bimanual manipulation increasingly depends on building pipelines that can collect large, high-quality datasets. Most existing approaches to data collection fall into two categories: \textbf{teleoperated} \cite{zhao2024aloha,fu2024mobile, aloha2team2024aloha2enhancedlowcost, liu2025factrforceattendingcurriculumtraining, dass2023patopolicyassistedteleoperation, park2024dexhubdartinternetscale} and \textbf{in-the-wild} \cite{chi2024diffusionpolicyvisuomotorpolicy, zhaxizhuoma2025fastumiscalablehardwareindependentuniversal, xu2025dexumiusinghumanhand, shafiullah2023bringing, tao2025dexwild}.  

Teleoperation enables high-fidelity demonstrations but requires expensive setup and operator training before consistent demonstrations can be produced \cite{zhao2024aloha, Connan2018AWU, fu2024mobile}. For example, ALOHA introduces a low-cost bi-manual leader and follower system \cite{aloha2team2024aloha2enhancedlowcost} and demonstrates robust bi-manual skills. The leader is composed of un-actuated linkages of the same kinematics as follower robot and is controlled by a human operator. Once available, data collected from such systems typically transfers well, since it is collected directly from the target robot. However, such systems typically require a large-footprint leader-follower setup deployed at the data collection site.  Moreover, teleoperated follower robots takes a significant amount of practice to get accustomed to the lack of proprioceptive feedback for the human operator \cite{impacthaptic}. 

By contrast, in-the-wild methods aim to bypass these barriers by collecting data on disembodied end-effectors. The DobbE system~\cite{shafiullah2023bringing} performed data collection with a single parallel-gripper that closely matched the real robot's embodiment. For bi-manual manipulation, the UMI system~\cite{chi2024universalmanipulationinterfaceinthewild} introduced a handheld paradigm with two parallel grippers and cameras, allowing users to record bi-manual demonstrations. Similarly, DeXwild~\cite{tao2025dexwild} collected data in-the-wild, but using using motion capture gloves and deploying on anthropomorphic robot hands. However, since no robot embodiment is present during collection, significant effort is still required to map these demonstrations onto real robots. Furthermore, demonstrations that prove unfeasible under the kinematic constraints of the complete robot must be discarded.

Recently, a third category has emerged: wearable bi-manual data collection systems \cite{fang2024airexo, fang2025dexop,devito_exoskeleton }. These systems aim to collect robot-relevant data without requiring direct access to a robot, instead using wearable hardware that mirrors the robot’s embodiment. By doing so, they reduce embodiment mismatch while retaining the scalability of human demonstrations. This combines the strengths of both prior paradigms: scalable data collection with minimal post-processing, and demonstrations that align closely with robot embodiment.

Examples of dual arm systems within this wearable category include AirExo \cite{fang2024airexo}, a 14-DOF wearable exoskeleton for bi-manual data collection, and the dual-arm teleoperation system in \cite{devito_exoskeleton}. Both consist of long kinematic chains that are fastened to the torso and are embedded with joint encoders. These systems designed to provide mobile demonstrations for traditional dual-arm robots, while mitigating the footprint of the system as much as possible. However, they also follow the framework of two highly articulated arms fastened to a fixed world frame, which increases complexity and footprint, while only being able to leverage a small portion of the overall systems workspace for dexterity. In this work, we introduce a new wearable system within this category, with a novel kinematic design, that is designed to take full advantage of the system's workspace. By reducing redundancy between the grippers, we achieve a significantly more compact and lightweight design, which makes the manipulator both wearable for data collection and mountable on a robot arm for large-reach policy execution.

\begin{figure}[t!]
\setlength{\tabcolsep}{1.0mm}
\centering
    \includegraphics[width=0.85\linewidth]{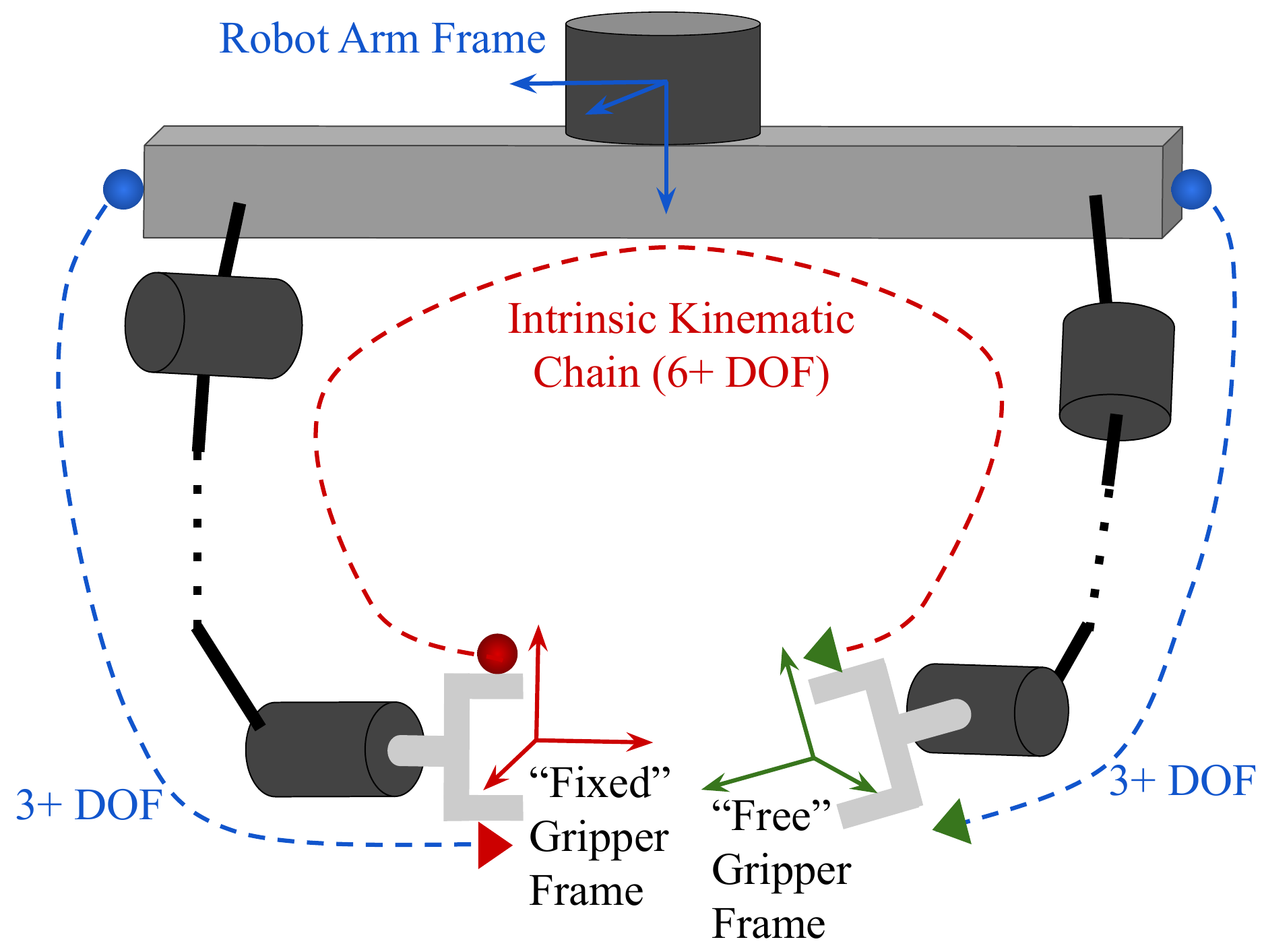}
\caption{\OURS concept. We use a bimanual system designed to mount on a larger robot arm. When designing the \OURS, we rely on a kinematic convention which treats one gripper as the base coordinate frame (the "fixed" gripper) and the other as the tip of the chain (the "free" gripper). Rather than computing each gripper’s pose relative to a robot tool frame (blue), we can then solve kinematic queries directly between the two grippers, relying on the fact that the complete kinematic chain connecting them (red) has sufficient DOFs for relative dexterity.}
\label{fig:minibe_concept}
\end{figure}

\begin{figure}[t!]
\setlength{\tabcolsep}{1.0mm}
\centering
    \includegraphics[width=\linewidth]{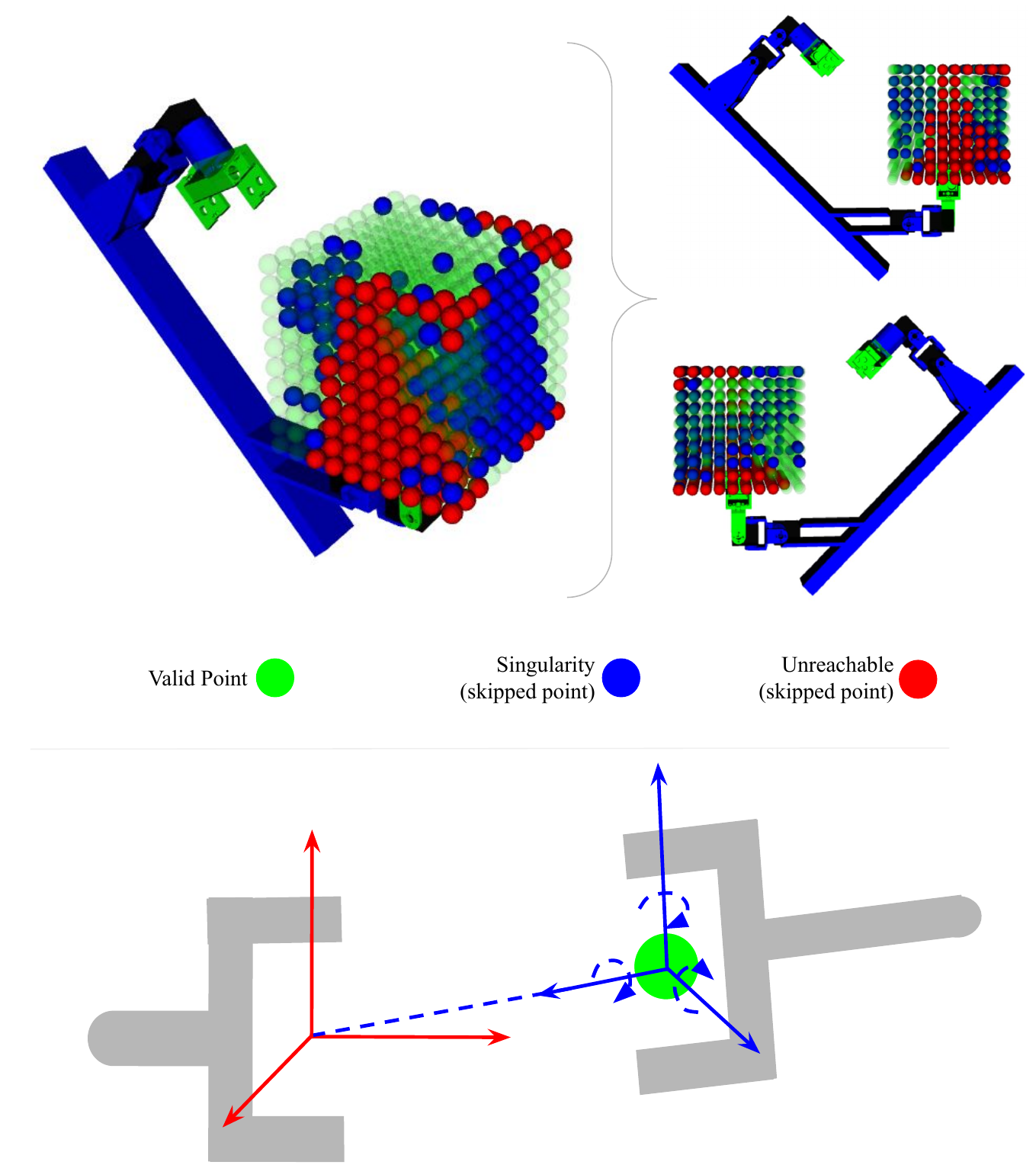}
\caption{We evaluate the kinematic dexterity between two grippers of a bi-manual kinematic chain. One of the grippers in the chain, dubbed the fixed gripper, serves as the base of our coordinate frame. We then set a desired volume for a workspace around the fixed gripper uniformly sample a set of points in this volume. We use each of these points as a desired position for the "free" gripper, oriented such that it points towards the fixed gripper (bottom image). If an IK solution for this query exists, obeys joint limits and self-collisions, and does not place the robot near a singularity, we consider the query point as reachable. The ratio of reachable to non-reachable points in the workspace is used as a metric for the relative kinematic dexterity between two grippers that are connected by a single kinematic chain.}

\label{fig:kd_metric_definition}
\end{figure}

\section{Design and Optimization}

The core idea of the \OURS is to realize a compact bimanual manipulator in which each arm individually has reduced mobility, but the combined kinematic chain formed between the two grippers retains sufficient articulation (6+ DOF) to achieve extensive relative mobility. In other words, through careful intrinsic design, the \OURS provides a dexterous workspace in which one gripper can be positioned and oriented with respect to the other with high flexibility. Absolute positioning in the environment can then be provided by mounting the system on a standard robotic arm, which then expands the kinematic chain for each gripper relative to the overall world frame. This concept is illustrated in Fig.~\ref{fig:minibe_concept}.

While 6+ intrinsic DOF are the theoretical minimum for arbitrary relative positioning, we find that the \emph{size} of the resulting dexterous workspace is strongly influenced by additional kinematic design choices, such as link lengths and joint limits, that will influence self-collisions and overall kinematic mobility at desired poses. . In this section, we describe our methodology for expanding this workspace while maintaining a low intrinsic DOF count, thereby keeping the \OURS compact and lightweight. In the following section, we show how these design properties enable efficient data collection and manipulation via imitation learning.

\begin{algorithm}[t]
\caption{Computation of the KD metric}
\label{alg:kdscore}
\begin{algorithmic}[1]
\Require Robot description (kinematics and joint limits), target workspace size and resolution, singularity threshold $\epsilon$
\Ensure Kinematic Dexterity metric KD $\in [0,1]$
\State Create a grid of points by sampling the target workspace at the desired resolution (see Fig.~\ref{fig:kd_metric_definition})
\State $N_{\text{total}} \gets$ total number of points in grid
\State $N_{\text{valid}} \gets 0$
\For{each point $\bm{p}_i$ in the grid}
    \State Compute free gripper pose $\bm{P}_i(\bm{p}_i)$ using Alg.~\ref{alg:poses}
    \State Robot pose $\bm{q}_i$=InverseKinematics($\bm{P}_i$)
    \If{no $\bm{q}_i$ exists }
        \State Skip this point (point is not reachable)
    \EndIf
    \State $\bm{J}_i$=Jacobian($\bm{q}_i$)
    \State Remove $\bm{J}_i$ columns corresp. to joints near limits
    \State Compute singular values of $\bm{J}_i$
    \If{smallest singular value $< \epsilon$}
        \State Skip this point (robot is near a singularity)
    \EndIf
    \State Count this point as valid: $N_{\text{valid}} \gets N_{\text{valid}} + 1$
\EndFor
\State KD $\gets$ $N_{\text{valid}}$ / $N_{\text{total}}$
\end{algorithmic}
\end{algorithm}

\begin{algorithm}[t]
\caption{Computation of free gripper poses}
\label{alg:poses}
\begin{algorithmic}[1]
\Require Workspace point $\bm{p}_i$
\Ensure Corresponding free gripper pose $\bm{P}_i$

\State Compute rotation $\bm{R}(\bm{p}_i)$ such that approach direction (z-axis) of free gripper points towards origin
\State$\bm{P}_i \gets \left(\bm{R}\left(\bm{p}_i\right), \bm{p}_i\right)$
\end{algorithmic}
\end{algorithm}

\subsection{Kinematic Analysis}

Conventional bimanual robots are described as two independent chains rooted at a shared base. Here, however, we focus exclusively on the \emph{intrinsic} DOF of the \OURS and their ability to control the relative pose of the two grippers. To this end, we adopt a different convention: the system is modeled as a single kinematic chain, with the base frame coincident with one gripper and the tool frame at the other gripper (Fig.~\ref{fig:minibe_concept}). 

To evaluate relative dexterity, we introduce the \textbf{Kinematic Dexterity (KD) metric}. Conceptually, the KD metric is designed to test the ability of one gripper to achieve a wide range of useful poses w.r.t. the other gripper.

The KD metric is defined as follows. One gripper, referred to as the "fixed" gripper, is treated as stationary and establishes the origin of the coordinate frame. The other gripper, referred to as the "free" gripper, is evaluated based on its ability to attain a diverse set of poses relative to this fixed frame. To this end, we uniformly sample candidate target poses, consisting of both translational and rotational components, within a predefined rectangular region centered on the fixed gripper. For each sampled pose, we employ standard inverse kinematics (IK) to assess feasibility, considering both joint-limit constraints and potential self-collisions. The KD metric is then defined as the proportion of sampled poses that are kinematically feasible under these conditions. The procedure is described in Alg. 1 and also illustrated in Fig. \ref{fig:kd_metric_definition}

We note that other formulations of dexterity metrics exist (e.g., manipulability ellipsoids, ergonomics-inspired indices~\cite{kd_metric1, kd_metric_2_ergo, kd_metric3}); however, we are interested here specifically in the relative positioning and orienting ability most relevant to bimanual tasks. For collision- and joint limit-aware IK, we use the KDL plugin~\cite{orocos_kdl} in the MoveIt library~\cite{coleman2014reducingbarrierentrycomplex}, which ensures realistic consideration of such constraints. 

\begin{figure*}[p!]
\centering
    \includegraphics[width=\linewidth]{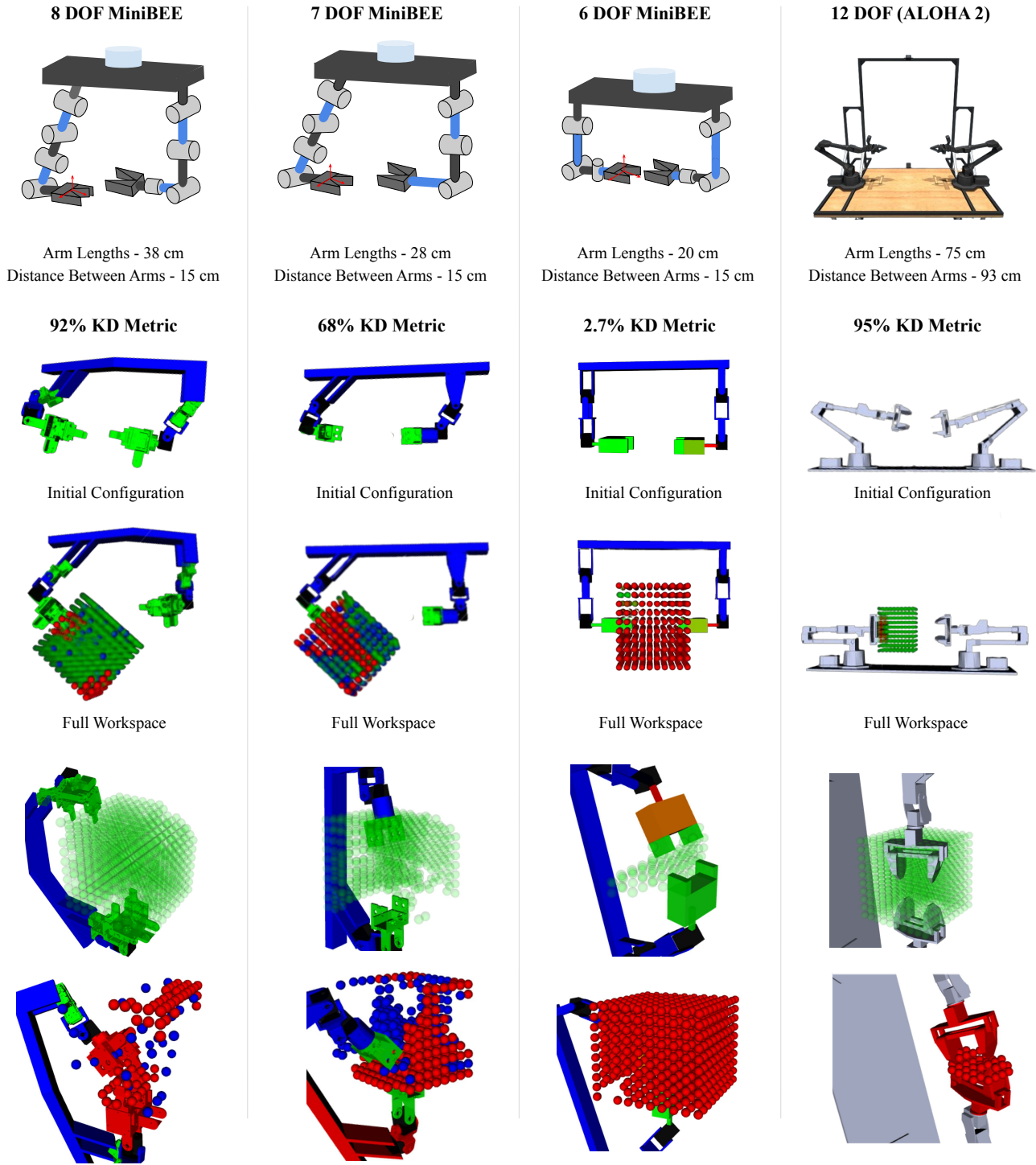}
\caption{Kinematic analysis for possible \OURS configurations as well as a state of the art bimanual system. For each configuration, we follow the procedure outlined in Algorithm 1 to compute the KD-metric for a desired fixed rectangular workspace with a side length of 20 cm. Successful desired points in the workspace are shown in green, points skipped due to IK solutions being near singularities are shown in blue, and points where there are no collision-free IK solutions are shown in red. The 8-DOF \OURS performed the best, as expected, and was comparable in performance to traditional bimanual systems that use two full mobility arms. Reducing the configuration to 7 and 6-DOF reduced the dexterity score significantly. For all experiments shown in the paper, we used the 8-DOF configuration, as it provided the best balance between complexity, size, and performance. The state of the art robot performs perfectly on our metric, but has 6 degrees of redundancy between the grippers. Overall, our KD metric correlates with the total degrees of freedom, and overall kinematic redundancy between the two gripper frames. }  
\label{fig:kd_comparison}
\end{figure*}

\subsection{Design Characterization}

With the KD metric established, we evaluate multiple possible \OURS configurations against each other, and also against an existing bimanual platform (Fig.~\ref{fig:kd_comparison}). We start with a configuration with 6 intrinsic DOFs, the theoretical minimum sufficient for relative dexterity. However, in the presence of realistic constraints such as self-collisions and joint limits, we find that adding more intrinsic DOFs progressively increases redundancy between the grippers and thus relative dexterity. We thus also test 7- and 8-DOF configurations of the \OURS using the same metric.

For each configuration, we used realistic joint limits and collision geometries based on available off-the-shelf servomotors (in our case, the Dynamixel X family). Of course, given a number of DOFs, many kinematic configurations are possible. Here, we selected the exact kinematics in each case empirically, based on design intuition. The KD metric allowed to quickly evaluate candidates, and select the best one in each class. A very compelling direction for future work would be to automate this process, and perform a full-fledged design optimization process with our KD metric as a fitness function.

While not directly comparable to the \OURS, we also compute the KD metric for an existing bimanual system, namely the ALOHA manipulator~\cite{zhao2024aloha}. We note that the ALOHA follower robot is not intended for wearability and kinesthetic teaching. However, with 12 total DOFs between the two arms, it represents a standard of performance in terms of relative dexterity. As such, it provides us with useful grounding information for the capabilities of the \OURS.

For each system, we applied Algorithm~\ref{alg:kdscore} with a fixed 20~cm cubic workspace, defined with one face centered at the fixed gripper and extending toward the free gripper. The same workspace size was used for all \OURS variants and the ALOHA system, despite the latter’s longer link lengths, to ensure a fair comparison.

Across \OURS variants, as expected, we find that more intrinsic DOF yields higher KD scores. The 8-DOF configuration maintained feasible kinematics over most of the workspace, with singularities appearing only when some of the joints become unusable to the robot as they hit their limits. Reducing to 7 or 6 DOFs sharply lowered performance: the 7-DOF system encountered frequent singularities and self-collisions, while the 6-DOF system often failed to find any solutions for many of the points. These results show that some redundancy between grippers is essential for avoiding self-collision, limiting singularities, and overall preserving dexterity. We therefore select the 8-DOF \OURS as it offers the best trade-off between compactness, complexity, and performance, while remaining competitive with more traditional bi-manual systems. 

The ALOHA system provides an interesting reference point. Its 12-DOF structure attains a perfect KD score. Although the 8-DOF MiniBEE achieves a slightly lower score, it is still comparable with fewer joints and a much smaller footprint overall. Based on this analysis, we believe that the \OURS provides relative dexterity comparable to state of the art system, while remaining compact and lightweight enough to be either worn for kinesthetic teaching or mounted on a robot arm for high reach, features that we put to use in the next section.

\begin{figure*}[t!]
\centering
    \includegraphics[width=\linewidth]{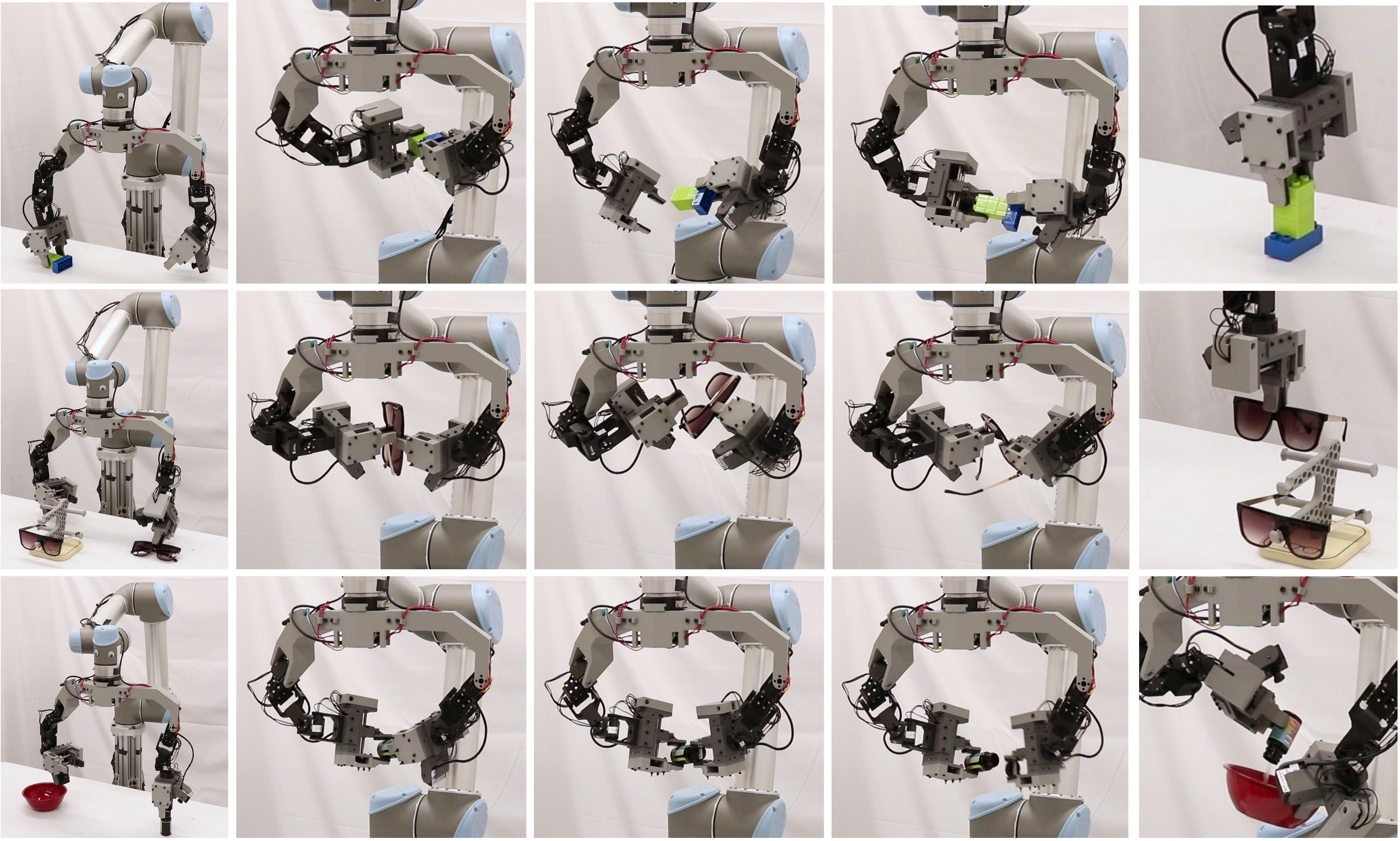}
\caption{We conduct a set of three manipulation experiments with \OURS. The first task, shown in the first row, is to grasp an object from the table, reorient it by handing it off to the other gripper, and place it on the table in a different pose. The second task, shown in the second row, is for \OURS to pick up a pair of folded sunglasses from the table, unfold them with the other gripper, and then place on a display rack. The third task is to pick up a closed pill bottle, unscrew the cap, pick up a bowl with the other gripper, and then empty the contents of the bottle into the bowl. \OURS performs these task with a high rate of success, and overall this set of experiments highlights the ability for our design to achieve bi-manual dexterity.}  
\label{fig:tasks}
\end{figure*}

In addition to kinematic capabilities, static payload capacity is an important mechanical characteristic. For our 8-DOF design used in the experiments (Fig. \ref{fig:tasks}), we calculated payload capacity for both the worst-case and a natural manipulation pose. We assumed that all motors operate at 30\% of stall torque given by the manufacturer, link masses are concentrated at the joints, and bracket/fastener masses are negligible. The worst-case pose yields a payload of 18g, including the 460g gripper mass, though this pose is unlikely during manipulation. In a more natural pose, with grippers facing each other and the UR5 tool frame mostly parallel to the floor, the payload increases to 168g in addition to the gripper mass. This is sufficient for our light-object, fine-manipulation tasks, though future iterations could improve capacity with lighter grippers or consideration of payload in our design analysis.

\section{Achieving Dexterous Manipulation}

Having described the overall design of the \OURS, we now turn to the goal of achieving dexterous bimanual manipulation. In particular, we rely on imitation learning from human demonstrations, a method recently shown to achieve skilled manipulation in this context~\cite{zhao2024aloha, chi2024universalmanipulationinterfaceinthewild, fang2024airexo, fu2024mobile}.

The fundamental advantage of \OURS, as described in the previous section, is that it enables intrinsic bimanual manipulation with fewer DOF than existing alternatives. It is compact, lightweight, and can be worn directly by an operator for kinesthetic data collection. This significantly reduces the difficulty of collecting demonstrations compared to systems with large robot arms, and even more so compared to leader-follower systems requiring four arms in total.

Compared to "in-the-wild" systems where the operator uses robot end-effectors, \OURS offers several advantages. First, the relative pose of the grippers is tracked through encoders in the kinematic chain, eliminating the need for external tracking or SLAM during intrinsic manipulation. Second, demonstrations are collected under the full set of kinematic constraints of the real system, ensuring they are feasible on hardware. Finally, \OURS is lightweight enough to fit within the payload of standard robot arms, extending reach for larger workspaces, unlike other wearable exoskeletons such as \cite{fang2024airexo}.

The self-tracking capabilities of \OURS apply during intrinsic manipulation, where the object is held by, or transferred between, the grippers. When manipulating an object fixed in an external reference frame, intrinsic encoders alone are insufficient. If one gripper remains fixed, intrinsic DOFs still suffice, but if both must move w.r.t. the external frame, additional tracking is needed, similar to in-the-wild systems. For this paper, our policies will focus on demonstrating the intrinsic dexterity between the two grippers, not the dexterity relative to the environment.

In this section, we describe our setup for data collection and policy training, and illustrate \OURS's performance on real-world dexterous bimanual tasks.

\subsection{Data Collection and Policy Training}

To collect data for policy learning, we leverage the fact that the same device can be used for data collection and policy deployment. The \OURS is worn and directly controlled by a human operator. A custom back brace (Fig.~\ref{fig:eyecandy}) supports the weight of the \OURS and stabilizes the system during demonstrations. Two ergonomic handles allow the operator to control the arms, and integrated triggers provide continuous control of the grippers, enabling natural modulation of grasp force and aperture throughout each task. The parallel grippers are equipped with RGB cameras, and the joints are driven by servos with embedded encoders to record joint angles. The total weight of \OURS, with the handles attached to the grippers, is 2.0kg and can be easily supported by our custom back brace, shown in Fig \ref{fig:eyecandy}, and also the UR5 robot arm. 

As mentioned earlier, in this paper, we focus on the intrinsic dexterity between the two grippers. Therefore, we do not collect information about the global positioning of the \OURS relative to an external fixed frame in the environment. As such, the components of a task where an object is first acquired from the environment, or placed back after task completion, are executed open loop. However, any bimanual component of the task is autonomous, using a policy based on the data collection procedure described here. 

Relative gripper poses are continuously self-tracked during demonstration collection, and all sensor signals are logged throughout the trajectory at 40 Hz. The observation space consists of RGB images from two cameras mounted on the grippers, as shown in Fig. \ref{fig:eyecandy}. The images are streamed at 640×480, then cropped to 480×360 to focus on the gripper and manipulated objects, and then downsampled to 320×240. The joint angles are recorded from the encoders embedded in the servos. The observation space also includes the relative pose between the two grippers during policy training and deployment. The action space comprises binary commands for gripper aperture and continuous commands for arm joint positions, directly derived from the operator’s handle inputs. This representation enables the policy to reproduce smooth, continuous manipulation behaviors rather than discrete open/close events.

To address the distribution shift between training and deployment—where handles are visible in demonstration observations but absent during robot execution—we additionally collect a small set of replayed trajectories, which are generated by the recorded data at 20 Hz. These provide handle-free visual inputs while preserving the underlying motion, helping the learned policy adapt to the deployment setting.

For policy training, we use Diffusion Policy (DP) \cite{chi2024diffusionpolicyvisuomotorpolicy}, which models the trajectory distribution with a conditional denoising diffusion process. In our implementation, we use a longer observation horizon for low-dimensional state inputs (e.g., gripper poses and joint angles) to better capture temporal dependencies, while restricting image inputs to the most recent frame for computational efficiency. The Diffusion Policy works at 5 Hz, which is trained by the down-sampled data.

\subsection{Experiments and Performance}

We used the method above on a total of 3 tasks, each of them requiring bimanual dexterity for completion. All tasks are illustrated in Fig.~\ref{fig:tasks}. For each task, there is a component where a single gripper must pick the object from a table in the environment. As stated, we are focusing only on demonstrating the intrinsic dexterity of our system, and therefore any pick up or placing of objects from the environment is a recorded trajectory. However, any manipulation between the grippers is done by a trained diffusion policy. 

\subsubsection{Task 1} Object re-orientation via double handoff. Here, the \OURS re-orients a grasped object by first transferring it to the other gripper, then transferring it back to the original gripper in a different pose. Such re-orientation is commonly needed in many applications, where the grasp originally available for an object is not the same one needed for later use.

\subsubsection{Task 2} Unfolding sunglasses. Here, the \OURS holds a folded pair of sunglasses in one gripper, then uses the other gripper to unfold the glasses, followed by placing the unfolded glasses on a shelf. This task highlight non-prehensile relative dexterity. 

\subsubsection{Task 3} Unscrewing and emptying bottle. Here, the \OURS uses both gripper to first unscrew a bottle, then to pour its contents into a bowl. This task highlights both prehensile and non-prehensile relative dexterity,

For each of these tasks, we collect a total of $N=45$ demonstrations via the wearable \OURS as described in the previous section. We then augment the training set with $M=5$ replay trajectories. We use these demonstrations to train autonomous policies for all the bimanual components of the tasks. Finally, we mount the \OURS on a robot arm, and test autonomous execution of the complete tasks. 

Our results show a 20/20 success rate for Task 1, 19/20 success for Task 2, and 18/20 success rate for Task 3. For Task 2, the lone failure was due to an imprecise grasp of the temples (or folding arms) of the glasses. For Task 3, both failures were due to incomplete lid unscrewing. Representative executions of all tasks are shown in Fig.~\ref{fig:tasks} as well as the Supplementary Video. Overall, these examples showcase the ability of the \OURS system to achieve dexterous bimanual manipulation based on demonstrations collected kinesthetically via a wearable system, leveraging its compact and lightweight nature.

\section{Conclusion and Future Work}

In this paper we have introduced \OURS, a concept for a new form factor for bimanual manipulation. \OURS attempts to combine the best features of both in-the-wild and leader-follower systems for data collection of bimanual manipulation. Like leader-follower systems, it comprises a two-gripper kinematic chain allowing fully self-tracked relative poses of the two grippers. However, through careful kinematic design analysis, \OURS uses a reduced articulation model, while still maintaining a level of dexterity (computed based on the relative positioning ability of the two grippers) comparable to much larger systems that use many more articulations.

This kinematic approach in turn enables a compact and lightweight design of the overall \OURS, which can be either worn directly by an operator (for kinesthetic data collection) of mounted on a robot arm (for high reach during autonomous policy execution). We demonstrate both of these modes by collecting kinesthetic data and training policies for multiple tasks requiring bimanual dexterity, both prehensile and non-prehensile. Our results show that the \OURS is indeed kinematically capable of such tasks, executing the policies trained from wearable data collection with high success rates.

While \OURS exhibits a high level of bimanual dexterity even in a compact package, additional areas of improvement remain. In particular, in future work we aim to implement self-tracking of uni-manual tasks executed relative to a fixed frame in the environment (possible in the case where the other gripper is kept fixed relative to the manipulation frame) in order to allow closed-loop policies for object acquisition from the environment. Other possible areas of improvement include better models of the \OURS position control system, to compensate for different gravity effects during data collection (grippers held by the operator) and policy execution (free-floating grippers). Finally, our KD metric used to assess different kinematics can be improved to include notions of torque, payload, and mass. Nevertheless, we believe \OURS is a novel approach to bimanual dexterity and a step towards highly dexterous bimanual systems that can easily deployed for a wide range of applications in human environments.

\bibliographystyle{IEEEtran}
\bibliography{references}

\addtolength{\textheight}{0cm}   

\end{document}